\documentclass{article}

\usepackage[preprint]{neurips_2022}

\usepackage[utf8]{inputenc} % allow utf-8 input
\usepackage[T1]{fontenc}    % use 8-bit T1 fonts
\usepackage{url}            % simple URL typesetting
\usepackage{booktabs}       % professional-quality tables
\usepackage{amsfonts}       % blackboard math symbols
\usepackage{nicefrac}       % compact symbols for 1/2, etc.
\usepackage{microtype}      % microtypography
\usepackage{xcolor}         % colors

\usepackage{color}
 \definecolor{mydarkblue}{rgb}{0,0.08,0.45}
\usepackage[colorlinks,citecolor=mydarkblue,urlcolor=mydarkblue,linkcolor=mydarkblue]{hyperref}
\usepackage{multirow}
\usepackage{makecell}
\usepackage{tabu}
\usepackage{graphicx}
\usepackage{subfigure}
\usepackage{amsmath}
\usepackage{enumitem}
\usepackage{capt-of}

\usepackage{cleveref}
\crefname{section}{\S}{\S\S}
\Crefname{section}{\S}{\S\S}
\crefname{table}{Table}{Tables}
\crefname{figure}{Figure}{Figures}
\crefname{algorithm}{Algorithm}{}
\crefname{equation}{eq.}{}
\crefname{appendix}{App.}{}
\crefformat{section}{\S#2#1#3}

\title{Rationale-Augmented Ensembles in Language Models}

\author{%
   Xuezhi Wang, Jason Wei, Dale Schuurmans, Quoc Le, Ed Chi, Denny Zhou \\
   Google Research, Brain Team \\
   \texttt{xuezhiw@google.com}\\
}

\begin{document}

\maketitle

\begin{abstract}

% Recent research has shown that rationales, or step-by-step chains of thoughts, can be used to improve tasks that require multi-step reasoning.
% However, existing prompting-based approaches are sensitive to prompt engineering and do not guarantee that the rationales used to prompt the model are optimal.
% In this paper, we study the role of \textit{rationales} in few-shot in-context learning, and show how we can reliably shift the paradigm from (input $\rightarrow$ output) to (input, \textit{rationale} $\rightarrow$ output) when prompting the language model.
% First, simply adding rationales might hurt task performance, and we identify the key reason as the sub-optimality of the rationales used in the prompts.
% Second, to overcome this sub-optimality, we unify the framework of rationale-augmented ensembles, and identify the key component as \textit{rationale sampling} in the output space.
% We show that the framework can extend to many common NLP tasks, even those that do not traditionally leverage intermediate steps (e.g., question answering, words-in-context, sentiment analysis), and it is a reliable way towards more accurate and interpretable natural language understanding: it outperforms both standard prompting without rationales, and rationale-based chain-of-thought prompting on most tasks; at the same time, it provides rationales for free to interpret models' predictions more easily.

Recent research has shown that \textit{rationales}, or step-by-step chains of thought, can be used to improve performance in multi-step reasoning tasks.
We reconsider rationale-augmented prompting for few-shot in-context learning, where (input $\rightarrow$ output) prompts are expanded to (input, \textit{rationale} $\rightarrow$ output) prompts.
For rationale-augmented prompting we demonstrate how existing approaches, which rely on manual prompt engineering, are subject to sub-optimal rationales that may harm performance.
To mitigate this brittleness, we propose a unified framework of \textit{rationale-augmented ensembles}, where we identify \textit{rationale sampling} in the \textit{output} space as the key component to robustly improve performance.
This framework is general and can easily be extended to common natural language processing tasks, even those that do not traditionally leverage intermediate steps, such as question answering, word sense disambiguation, and sentiment analysis. 
We demonstrate that rationale-augmented ensembles achieve more accurate and interpretable results than existing prompting approaches---including standard prompting without rationales and rationale-based chain-of-thought prompting---while simultaneously improving interpretability of model predictions through the associated rationales.

\end{abstract}

\section{Introduction}

Recent progress on improving few-shot in-context learning in pretrained large language models has been achieved by expanding prompt exemplars with rationales, delivering successes in a variety of natural language reasoning tasks 
\citep{wei2022chain,palm,explanation_deepmind,selection_inference, step_by_step, maieutic_prompting}. 
These prompting-based approaches typically adopt manually-written rationales and therefore rely on the quality of prompt engineering, which usually does not ensure optimal rationales are provided for a given task.
Previous work has also shown that ``rationales'' can be useful 
for \textit{supervised learning}
in natural language tasks when added in the training data \citep{zaidan-etal-2007-using,ling-etal-2017-program,cobbe2021training,star}, but it remains unclear whether such rationales can be reliably useful in few-shot in-context learning \citep{ye2022unreliability}.

In this paper, we investigate the role of \textit{rationales} in few-shot in-context learning by conducting a systematic study over a wide range of NLP tasks. 
In particular, we seek to answer the following questions: 
(1) Why do rationales sometimes hurt task performance in few-shot learning?
and (2) How can one reliably leverage rationales in few-shot learning for general natural language tasks? 

\begin{figure}[t]
\vspace{-0.1in}
    \centering
    \includegraphics[width=\linewidth]{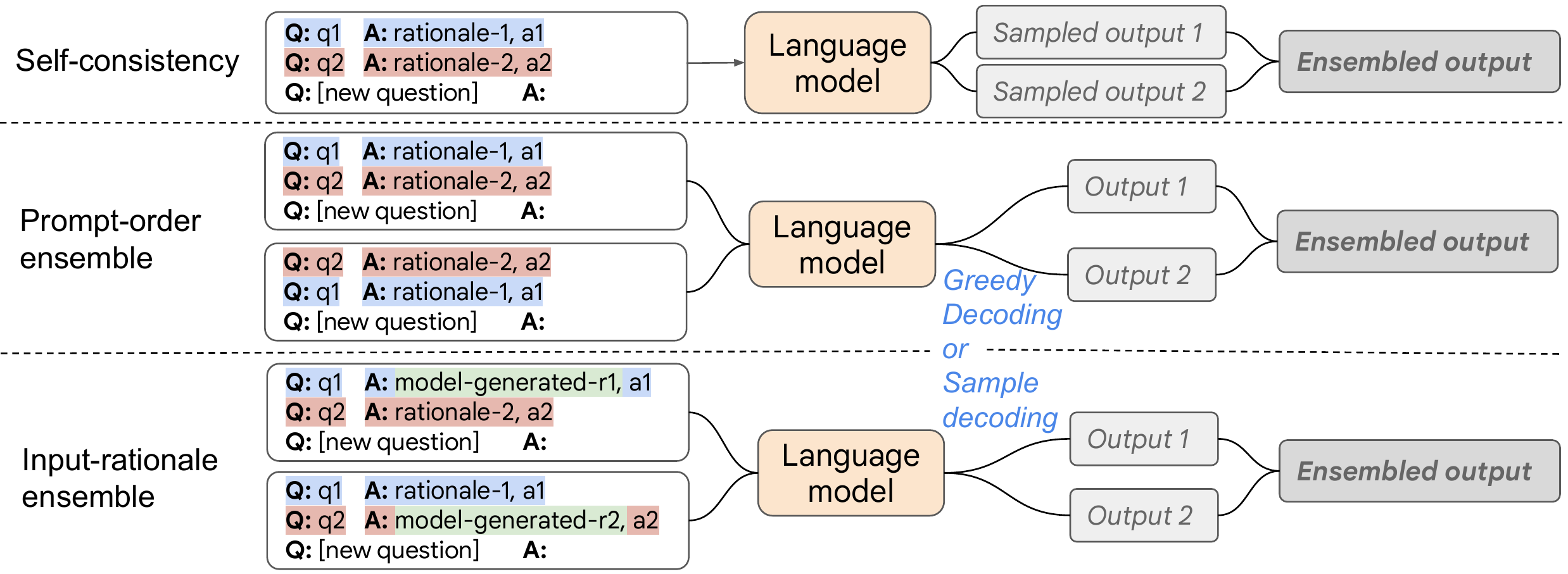}
    \vspace{-0.22in}
    \caption{An overview of different ways of composing rationale-augmented ensembles, depending on how the randomness of rationales is introduced. Here $q$, $a$, $r$ correspond to question, answer, and rationale, respectively. Rationales are human-written unless specified as model-generated.}
    \label{fig:rae_overview}
    \vspace{-0.08in}
\end{figure}

Below we show that, when shifting from the simpler paradigm of (input $\rightarrow$ output) prompts to expanded (input, \textit{rationale} $\rightarrow$ output) prompts, there is indeed a large variance in final task performance for few-shot in-context learning.
We identify the primary source of sensitivity as the \textit{sub-optimality} of the rationales used for prompting.
To overcome such sub-optimality, we develop a unified framework of \textbf{rationale-augmented ensembles}, where the idea is to aggregate over multiple rationales generated from the language model to reduce the brittleness of the results.
Ensemble aggregation can be achieved in a few different ways depending on how randomness over the rationales is introduced in the input or the output space, including
(1) self-consistency, 
where existing work \citep{self_consistency} has shown that task performance can be improved by sampling multiple language model outputs for ensembling,
(2) prompt-order ensembling, where previous work \citep{Lu2021FantasticallyOP,pmlr-v139-zhao21c} has shown that task performance is sensitive to the order of the exemplars in the prompts, and
(3) input-rationale ensembling, where human-written rationales can be replaced by model-generated rationales, leveraging the ability of language models to generate high-quality explanations \citep{wiegreffe2021reframing}.
Figure~\ref{fig:rae_overview} provides an overview of rationale-augmented ensembling approaches.

A key finding of this study is that \textit{rationale sampling} in the \textit{output} space is a central aspect of rationale-augmented ensembles contributing to their success.
That is, regardless of how the input or the prompt vary, task performance is best improved when sufficient diversity is introduced by sampling rationales from the language model's decoder.
We also find that rationale-augmented ensembles reliably outperform existing rationale-based few-shot and zero-shot prompting methods \citep{wei2022chain,step_by_step} across a variety of natural language processing tasks. 
Moreover, in cases where human-written rationales hurt task performance due to the sub-optimality of the rationales, rationale-augmented ensembling is able to fill the gap and reliably outperform standard few-shot prompting \citep{brown2020language} on most tasks.

Perhaps surprisingly, we also find that the proposed framework can be used to improve few-shot learning in common natural language processing tasks, even including tasks where explicit intermediate steps might not be necessary, such as question answering \citep[BoolQ;][]{clark2019boolq}, word sense disambiguation \citep[WiC;][]{pilehvar-camacho-collados-2019-wic}, sentiment analysis \citep[SST-2;][]{socher-etal-2013-recursive}, and paraphrase identification \citep[QQP;][]{WinNT}. 
We conjecture that, in principle, any natural language processing task can be usefully augmented with ``rationales'' that represent the thought processes needed to achieve accurate and interpretable results in few-shot in-context learning.

Existing work on interpretability usually focuses on improving the explanation of model predictions via supervised learning, which requires large amounts of human labeled explanations to be collected \citep{zaidan-etal-2007-using,esnli,rajani-etal-2019-explain,wt5}, while remaining agnostic to improving final task performance.
In contrast, we show that the framework proposed in this paper can leverage very few human-written rationales (as $K$-shot exemplars where $K$ is usually very small, e.g., 3 to 6) and still generate ensembles that can improve task performance significantly. 
The proposed framework does not require additional fine-tuning \citep{thoppilan2022lamda,star}, verifiers \citep{cobbe2021training}, calibrators \citep{ye2022unreliability}, or any use of an auxiliary dataset \citep{star,better_reasoner}, making it applicable to any off-the-shelf large language model.
As a general approach to obtaining more accurate and more interpretable natural language understanding, rationale-augmented ensembles also provide more accurate assessments of the performance gains contributed by rationales in few-shot in-context learning.

\section{Rationale-Augmented Ensembles in Language Models}
% The key idea of rationale-augmented ensemble is to overcome the sub-optimality of human-written rationales in few-shot in-context learning by sampling from a large amount of language-model-generated rationales.
% This is based on the observation that large language models are good few-shot reasoners \citep{scratchpad,wei2022chain}, and that language models can generate high-quality explanations \citep{wiegreffe2021reframing}.

We investigate the role of rationales in few-shot in-context learning, first interrogating the sensitivity of final performance to rationale quality, then developing a unified perspective on rationale-augmented ensembles that seek to reduce sensitivity and improve final performance.

% We first conduct a study showing that human-written rationales might not be optimal for the final task performance on a range of NLP tasks, including e-SNLI \citep{esnli}, BoolQ \citep{clark2019boolq}, WiC \citep{pilehvar-camacho-collados-2019-wic}, and SST-2 \citep{socher-etal-2013-recursive}, as some performance variances have been observed in  \cite{wei2022chain,ye2022unreliability}.

\subsection{Optimality of the rationales in few-shot learning.}
\label{sec:optimality}
Given that rationale-augmented prompting has been shown to exhibit variable performance \citep{wei2022chain,ye2022unreliability}, we first investigate the sensitivity of task performance to rationale quality across a range of natural language tasks, including e-SNLI \citep{esnli}, BoolQ \citep{clark2019boolq}, WiC \citep{pilehvar-camacho-collados-2019-wic}, and SST-2 \citep{socher-etal-2013-recursive}, finding that human-generated rationales can indeed be sub-optimal.

% For each of the tasks, we pick $K$ (3 to 6) exemplars from the training set, and manually write a set of rationales for each of the exemplars, then we use them as seeds to further sample additional rationales from the language model: we leave one question from the exemplars out, and use the rest of the exemplars with human-written rationales as prompts.
% Specifically, we sample 1,024 rationales for each exemplar and only keep rationales where the final answer is consistent with the ground truth answer.
% We compose new prompts in the following way: for each of the $K$ exemplars, we replace its human-written rationale with a randomly sampled generated-rationale, and keep the rationales of the other $K-1$ exemplars fixed. We repeat this for every exemplar and report the final task performance using these new prompts in Figure~\ref{fig:sampled_performance} (denoted as sampled-r-$k$ if the $k$-th rationale is replaced).

\begin{figure}[h]
\vspace{-0.1in}
    \centering
    \hspace{-0.1in}
    \includegraphics[width=0.5\linewidth]{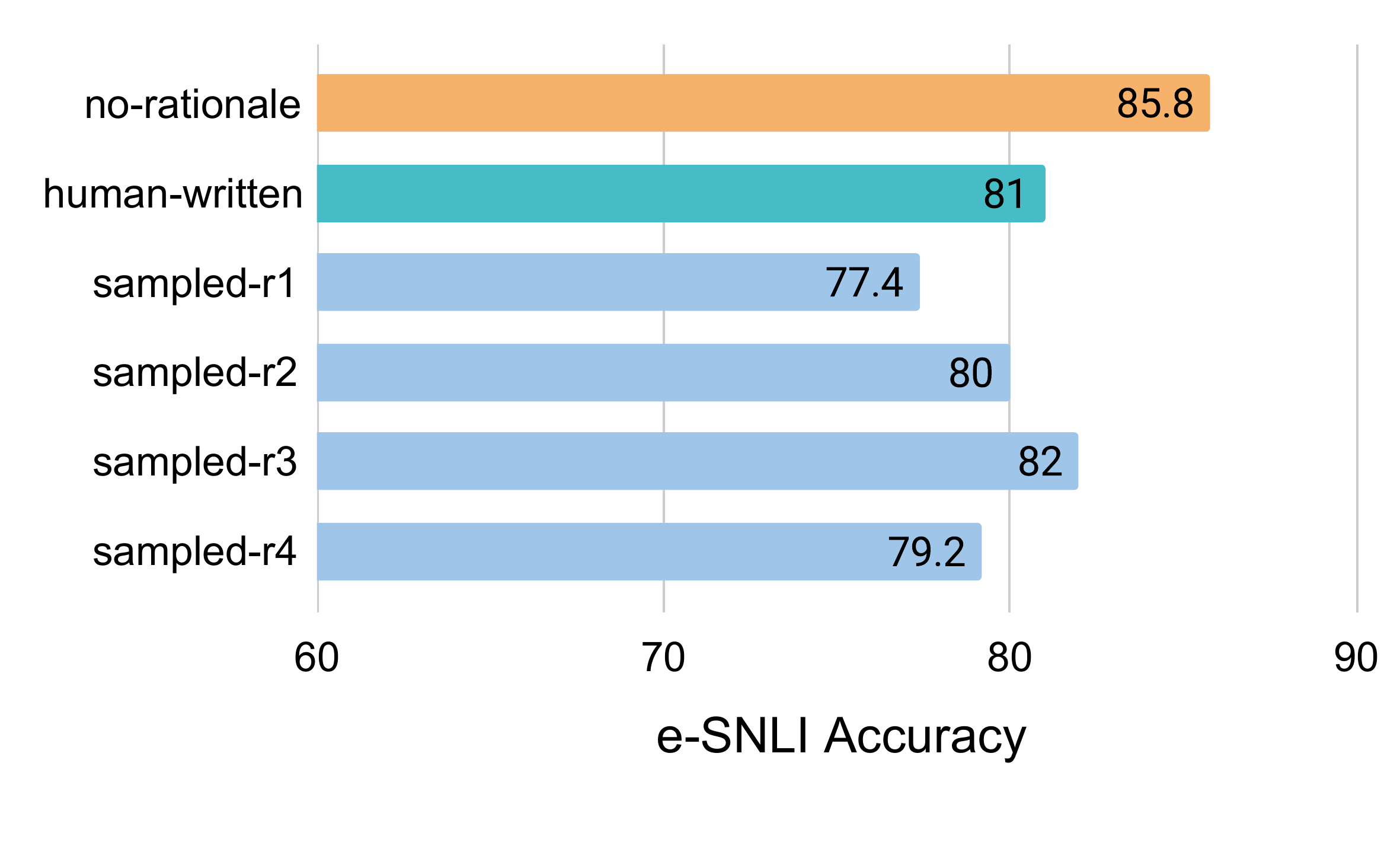}
    \hspace{-0.1in}
    \includegraphics[width=0.5\linewidth]{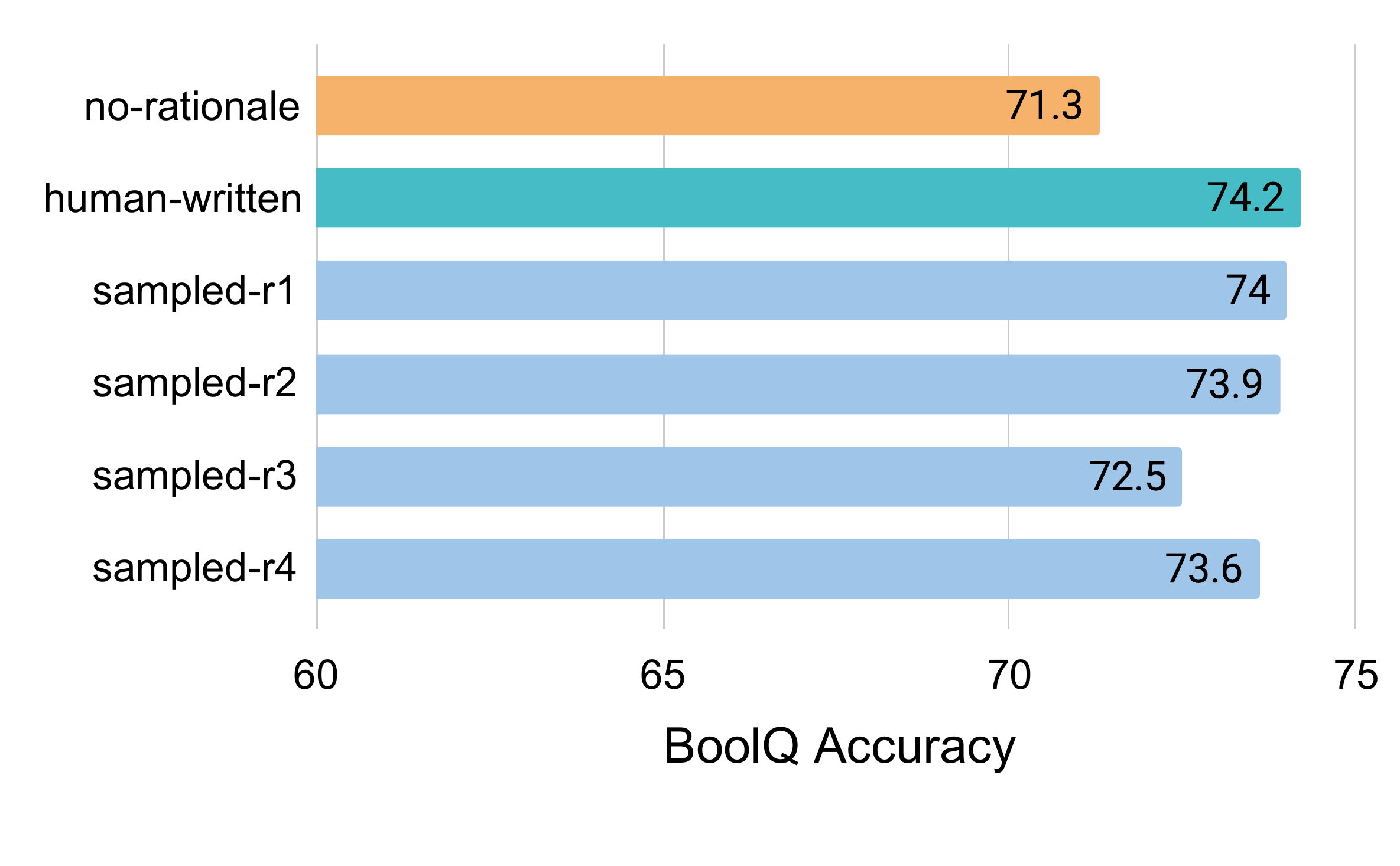}\\
    \vspace{-0.15in}
    \hspace{-0.1in}
    \includegraphics[width=0.5\linewidth]{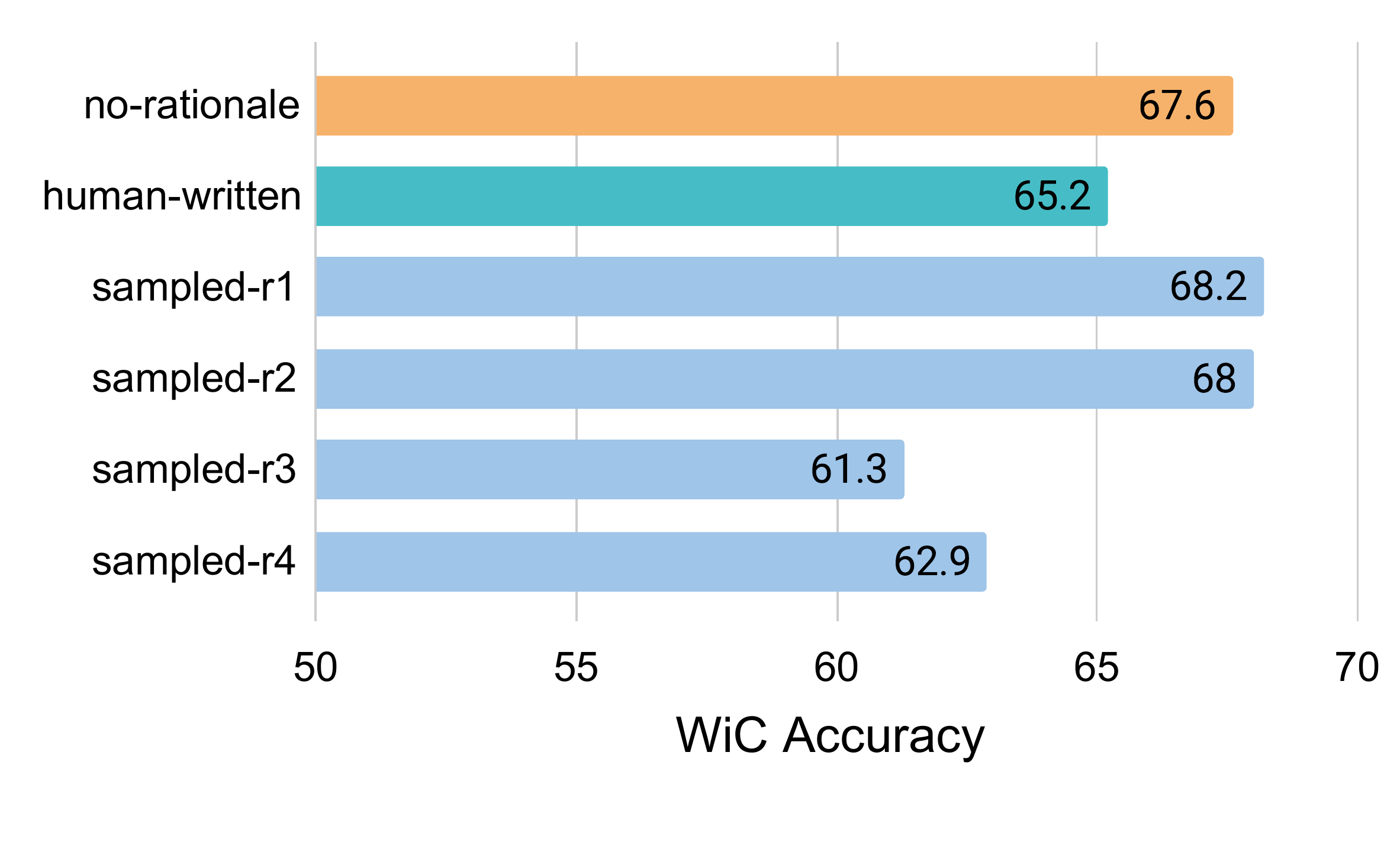}
    \hspace{-0.1in}
    \includegraphics[width=0.5\linewidth]{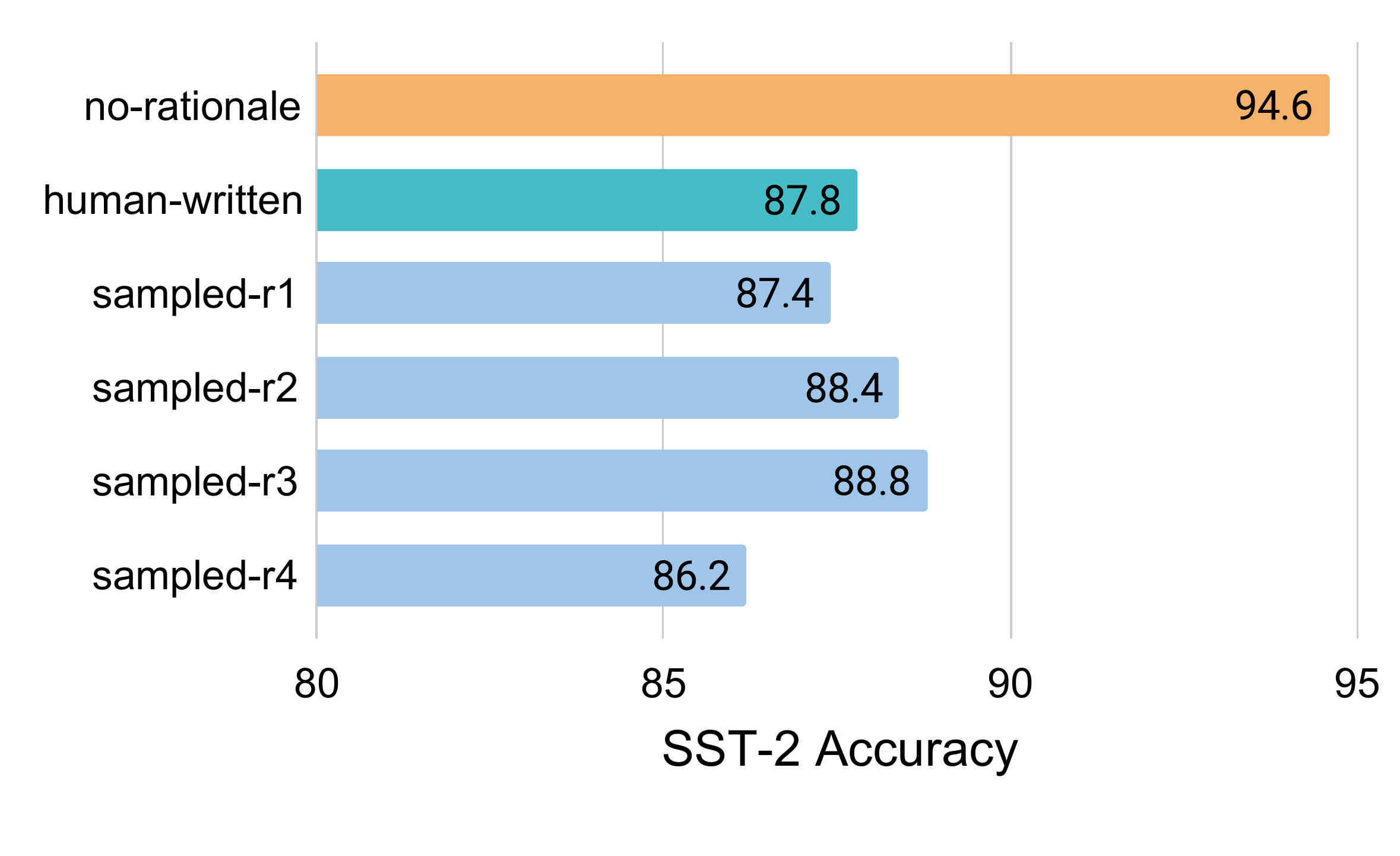}
    \hspace{-0.1in}
    \vspace{-0.2in}
    \caption{The performance varies depending on which rationales are used in the prompts for few-shot in-context learning. The exemplars in the prompts are exactly the same, only the rationales differ. The performance is evaluated with accuracy on the greedy decoded output using PaLM-540B \citep{palm}.}
    \label{fig:sampled_performance}
\end{figure}

For each task, we choose $K$ (4 to 6) exemplars from the training set, manually produce a set of rationales for each exemplar, then use these as seeds to generate additional rationales from the language model: we leave one question from the exemplars out, and use the rest of the exemplars with human-written rationales as prompts, then we can sample from the language model's decoder to obtain a large number of generated rationales for this question.\footnote{Specifically, we sample 1,024 rationales for each exemplar and only keep those where the final answer is consistent with the ground truth answer.}
Each new prompt is then composed as follows: for each of the $K$ exemplars, we replace its human-written rationale with a random sample from the generated rationales, while keeping the rationales of the other $K-1$ exemplars fixed. We repeat this for every exemplar and report the final task performance using the new prompts in Figure~\ref{fig:sampled_performance} (denoted as sampled-r-$k$ if the $k$-th rationale is replaced).

% First, we observe that compared with standard prompting (``no-rationale''), adding human-written rationales in the prompts may or may not achieve a better performance.
% In addition, the sampled rationales yield performances with a fairly large variance, showing that the quality of the rationales in the prompts does have a large effect to the final performance. In many cases the sampled rationales achieve a better performance compared to the human-written ones, indicating that the human-written rationales are usually not ``optimal'' in terms of task performance.
% Table~\ref{tab:example_rationale} shows examples of human-written rationales and model-generated rationales given the same question for common NLP tasks. We see that the model can learn to generate more diverse rationales using its pre-trained knowledge.

First, one can observe that compared with standard few-shot prompting (``no-rationale''), the addition of human-written rationales does not always yield better performances.
Moreover, the performance induced by sampled-rationales exhibits substantial variance, implying that the quality of the rationales in the prompts has a significant effect on final performance.
Often the sampled rationales exhibit better performance than the human-written ones, indicating that manually provided rationales can be far from ``optimal'' in terms of task performance.
Table~\ref{tab:example_rationale} shows examples of human-written rationales and two model-generated rationales for the same question, demonstrating that the model is able to generate diverse but reasonable rationales using its pre-trained knowledge.

\begin{table}[h]
\vspace{-0.1in}
\small
\centering
    \begin{tabu} to 1\textwidth { X[l] }
        % Task / Question & Human-written rationale & Model-generated rationale  \\
        \toprule
        \textbf{[e-SNLI]}  \textit{Premise: ``Children smiling and waving at camera." Based on this premise, can we conclude the hypothesis ``The kids are frowning." is true?}\\
        \textbf{Human-written rationale (crowd-sourced)}: One cannot be smiling and frowning at the same time.\\
        \textbf{Model-generated rationale}: \textcolor{blue}{Smiling and waving is not frowning}. $\mid\mid$ \textcolor{teal}{'Smiling' and 'frowning' are opposites.}
        \\
         \midrule
        \textbf{[BoolQ]}  \textit{can you use oyster card at epsom station?}\\
         \textbf{Human-written rationale}: Epsom railway station serves the town of Epsom in Surrey and is not in the London Oyster card zone.\\
         \textbf{Model-generated rationale}: \textcolor{blue}{Epsom Station is not currently a part of the Oyster Card Network.}  $\mid\mid$ \textcolor{teal}{'Oyster card' is a pre-pay card for public transport in London. Epsom Station is in Surrey}. \\
        % \midrule
        % \textbf{[HotpotQA]}  \textit{The Oberoi family is part of a hotel company that has a head office in what city?}\\
        %  \textbf{Human-written rationale}: The Oberoi family is part of the hotel company called The Oberoi Group. The Oberoi Group has its head office in Delhi.\\
        %  \textbf{Model-generated rationales}: \textcolor{blue}{The Oberoi family runs a chain of luxury hotels. The chain is called the Oberoi Hotels. The Oberoi Hotels have its head office in Delhi}. $\mid\mid$ \textcolor{teal}{The Oberoi family is part of the Oberoi Hotels and Resorts whose head office is in Delhi, India}.\\
         \midrule
        \textbf{[WiC]}  \textit{Approach a task. To approach the city. Q: Is the word  ``approach" used in the same way in the two sentences above?}\\
         \textbf{Human-written rationale}: The first ``approach" means ``deal with", the second ``approach" means ``come near". \\
         \textbf{Model-generated rationale}: \textcolor{blue}{1) To tackle or deal with a task. 2) To come near}. $\mid\mid$ \textcolor{teal}{'To approach the city' means ``to get close to the city". 'Approach a task' means ``to deal with a task"}.\\
         \midrule
         \textbf{[QQP]}  \textit{Is the following question ``What can one do after MBBS?" the same as ``What do i do after my MBBS?"}\\
         \textbf{Human-written rationale}: Both are asking what can a person do after MBBS. \\
         \textbf{Model-generated rationale}: \textcolor{blue}{`One' means the same as `I'}. $\mid\mid$ \textcolor{teal}{'What can one do' and 'what do I do' mean the same thing in the context of the question}.\\
         \midrule
         \textbf{[SST-2]}  \textit{What is the sentiment of the following sentence? ``contains no wit , only labored gags"}\\
         \textbf{Human-written rationale}: ``contains no wit" is clearly a negative sentiment.\\
         \textbf{Model-generated rationale}: \textcolor{blue}{``no wit" and ``labored gags" both mean negative sentiments}.  $\mid\mid$ \textcolor{teal}{'no wit' indicates something negative and ``only labored gags" is also negative}.\\
         \bottomrule
    \end{tabu}
    \vspace{0.02in}
    \caption{Examples of human-written rationale and model-generated rationales from PaLM-540B \citep{palm}, representing multiple ways to reason about the same question. These examples show that many common NLP tasks can be augmented with rationales, even when the task itself does not require explicit intermediate steps.}
    \label{tab:example_rationale}
\vspace{-0.1in}
\end{table}

\paragraph{Simply including a rationale does not always improve task performance.}
From Figure~\ref{fig:sampled_performance} one can also see that, due to the sub-optimality of the rationales used, task performance can degrade when rationales are added to prompts in few-shot in-context learning. 
For example, on e-SNLI, WiC and SST-2, the performance achieved by written-rationales is significantly worse than standard few-shot prompting without rationales, consistent with the findings in \citep{ye2022unreliability}.

\subsection{Rationale-augmented ensembles}
Given that determining ``optimal'' rationales for few-shot in-context learning is difficult,%
\footnote{A line of existing work uses a train/validation set to determine the optimal prompts (either discrete or continuous), e.g., \cite{lester-etal-2021-power,gao-etal-2021-making}.
Such a setting is closer to fine-tuning rather than few-shot learning, due to the use of an additional  dataset for performance validation.}
it is natural to consider the use of \textbf{rationale-augmented ensembles} that can automatically aggregate across diverse rationales to overcome the brittleness of performance to sub-optimal human-written rationales.

\begin{table}[]
\setlength\tabcolsep{3pt}
    \centering
    \begin{tabular}{c|c|c}
    \toprule
        Rationale-augmented ensembles & Input/Prompt & Output \\
         \midrule
        Self-consistency \citep{self_consistency} & fixed & sampled \\ 
        Prompt-order ensemble \citep{Lu2021FantasticallyOP,pmlr-v139-zhao21c}  & shuffled & greedy/sampled \\
        Input-rationale ensemble, adapted from \citep{wiegreffe2021reframing} & sampled & greedy/sampled\\
        \bottomrule
    \end{tabular}
    \vspace{0.02in}
    \caption{Methods for generating rationale-augmented ensembles in language models.}
    \label{tab:overview}
    \vspace{-0.2in}
\end{table}

We define a rationale-augmented ensemble as introducing an additional latent variable (the ``rationales'') that can be sampled and ultimately marginalized out (see Figure~\ref{fig:rae_overview} for examples).
Depending on the stage where the sampling occurs, the approaches to rationale ensembling can be categorized as follows (summarized in Table~\ref{tab:overview}):
\begin{itemize}[leftmargin=0.8cm]
    \item Self-consistency \citep{self_consistency}, where the input/prompt is fixed, and multiple rationales are sampled from the language model's decoder.
    \item Prompt-order ensemble: Given that task performance has been observed to be sensitive to prompt ordering \citep{Lu2021FantasticallyOP,pmlr-v139-zhao21c}, the order of exemplars in prompts can be permuted to elicit multiple rationales in the decoder.
    \item Input-rationale ensemble: Leveraging the ability of large language models to generate high-quality explanations \citep{wiegreffe2021reframing}, model-generated rationales can replace human-written rationales in the input prompts (e.g., via the process described in Section~\ref{sec:optimality}), which can then be used to elicit multiple rationales in the decoder.
\end{itemize}
For each of these ensembling approaches, the model couples the generation of rationales and answers before taking a majority vote (more precisely, a plurality vote) to produce the final ensemble answer.
For both prompt-order ensembling and input-rationale ensembling, since the randomness is introduced in the \textit{input} space, one can either decode an output greedily with a rationale, or sample an output with a rationale in the \textit{output} space for each new prompt.
Interestingly, below we find that \textit{rationale sampling} in the \textit{output} space is the most important component in the overall rationale-augmented ensemble framework.
In particular, regardless of how the input/prompt varies, sampling in the output space is the key to achieving better task performance across a variety of natural language processing tasks. With this key component, we find that rationale-ensembling can significantly improve results over both standard prompting \citep{brown2020language} and rationale-based prompting \citep{wei2022chain,step_by_step} on common NLP tasks; the framework also provides rationales at no additional cost that can be used to better interpret model predictions.

\section{Experiments}

We conducted a series of experiments to compare the performance of rationale-augmented ensembles against existing approaches, across a variety of natural language processing tasks.
Overall, the results demonstrate that rationale-augmented ensembles can robustly improve task performance across alternative language models and model scales.

\subsection{Experiment setup}

\paragraph{Tasks and datasets.}
We considered a set of natural language tasks from GLUE \citep{wang-etal-2018-glue}, SuperGLUE \citep{superglue}, and other natural language processing benchmarks.
These tasks can be categorized as follows:%
\footnote{We use the test split for all tasks if the test split is available and has labels for evaluation, otherwise we use the dev split. Specifically, test split: ANLI, e-SNLI, OpenBookQA, ARC; dev/validation split: MNLI, RTE, BoolQ, Hotpot-QA, WiC, SST-2, QQP.
In addition, some of the datasets are too large to run large language models on, so we used the first 1,000 data points for HotpotQA, e-SNLI, MNLI, and QQP for evaluation.}
\begin{itemize}[leftmargin=0.6cm]
\vspace{-0.05in}
\item \textbf{Question Answering}:
For question answering, we include BoolQ \citep{clark2019boolq}, HotpotQA \citep{yang-etal-2018-hotpotqa}, and OpenBookQA \citep{openbookqa}.
\item \textbf{Natural Language Inference}:
For these tasks, we include ANLI \citep{nie-etal-2020-adversarial} with the three subsets (R1, R2, R3), e-SNLI \citep{esnli}, MNLI (matched/mis-matched) \citep{mnli}, and RTE \citep{dagan2005pascal,bar2006second,giampiccolo2007third,bentivogli2009fifth}.
\item \textbf{Word Sense Disambiguation}:
Here we use Word-in-Context  \citep[WiC;][]{pilehvar-camacho-collados-2019-wic}.
\item \textbf{Sentiment Analysis}:
we use the Stanford Sentiment Treebank v2  \citep[SST-2;][]{socher-etal-2013-recursive}.
\item \textbf{Paraphrase Identification}:
Here we use Quora Question Pairs \citep[QQP;][]{WinNT}.
\item \textbf{Reasoning}.
For reasoning tasks, we consider the AI2 Reasoning Challenge (ARC) \citep{Clark2018ThinkYH} for open-domain question answering with commonsense reasoning, as well as the grade-school math problems \citep[GSM8K;][]{cobbe2021training} for arithmetic reasoning.
\end{itemize}

\vspace{-0.15in}
\paragraph{Language models and prompts.}
To investigate whether rationale-augmented ensembles can robustly improve performance across language models, we evaluated the framework with two dense left-to-right, decoder-only transformer language models with varying scale: 
(1) PaLM-540B, a language model with 540-billion parameters \citep{palm}
and
(2) the public GPT-3 model with 175-billion parameters \citep{brown2020language,instructGPT}.

All experiments are conducted in the few-shot setting except the zero-shot CoT baseline \citep{step_by_step}, without any fine-tuning. 
For each task, we randomly choose $K$ examples from the training set as $K$-shot prompts, while maintaining a balanced label distribution and manually providing a set of rationales as the initial prompts; see Appendix~\ref{sec:appendix-prompt} for the full set of initial prompts and rationales used in each experiment.
We use the exact same exemplars in the few-shot prompts for all baselines and rationale-augmented ensembles. For standard few-shot prompting we omit the rationales.

\vspace{-0.1in}
\paragraph{Parameter settings.} Across all tasks, each rationale-augmented ensemble is generated by ensembling $m=40$ outputs from the language model.
For sampling in the language model, we use temperature sampling \citep{ACKLEY1985147,ficler-goldberg-2017-controlling} with temperature $T=0.7$.
The maximum number of decoded steps is set to $128$ in every case, except for GSM8K where we use $256$ to accommodate longer rationales needed to express extended reasoning chains.

\subsection{Results}

The results for the PaLM-540B model are shown in Table~\ref{tab:performance_1}, Table~\ref{tab:performance_2} and Table~\ref{tab:performance_3}, and give a comparison to two baseline approaches:
(1) standard few-shot prompting without rationales \citep{brown2020language}, and
(2) rationale-based prompting, including few-shot chain-of-thought (CoT) prompting \citep{wei2022chain}, and zero-shot CoT \citep{step_by_step} where the model is prompted with ``Let's think step by step'' to generate initial rationales then prompted with ``Therefore, the answer is'' to obtain the final answer.%
\footnote{We have found the zero-shot CoT approach yields slightly less controlled responses compared to few-shot based approaches, i.e., the model is less likely to generate a desired fixed answer like ``yes/no'', ``(a)-(e)'' even when we add guided prompts like ``The answer (yes or no) is'', ``among options (a) through (e)''.}

For each of the rationale-augmented ensembles, we specify the inputs as ``fixed", ``shuffled" (for prompt-order ensemble), or ``sampled" (for input-rationale ensemble); and the outputs as ``greedy" or ``sampled" depending on whether we decode the outputs greedily or sample the outputs from the language model's decoder.
Based on the results shown in the tables, a few key observations follow:
\begin{itemize}[leftmargin=0.6cm]
%\vspace{-0.1in}
    \item For each rationale-augmented ensemble strategy, the ``output-sampled'' version yields better final performance than the ``output-greedy'' version for almost every task.
This remains true regardless of whether randomness is introduced in the input space (i.e., whether the exemplars are shuffled in a prompt-order ensemble, or whether rationales in the exemplars are sampled in an input-rationale ensemble).
Although self-consistency has an ``output-sampled'' only version, given that the input/prompt is fixed, it also achieves comparable performance to the ``output-sampled'' versions of the other ensembling approaches.
These findings indicate that \textit{rationale sampling} in the \textit{output} space is the critical component for improving task performance, more so than the specific ensembling method used.
    \item The ``output-sampled'' version of each rationale-ensembling method almost always improves performance over standard prompting \citep{brown2020language} without rationales, as well as rationale-based few-shot and zero-shot prompting \citep{wei2022chain,step_by_step}.
There are a few exceptions, including MNLI-m/mm, SST-2, and QQP, from GLUE \citep{wang-etal-2018-glue}, where standard-prompting still exhibits the best performance.
We conjecture that the questions and answers in these tasks already appear frequently in the pre-training corpus, which allows simple memorization to perform well, whereas forcing the model to additionally provide rationales slightly degrades performance.
    \item Simply adding rationales as in \citep{wei2022chain,step_by_step} can sometimes degrade task performance compared to standard prompting (also observed in \citep{ye2022unreliability}), but rationale-augmented ensembling reliably boosts performance beyond both rationale-based and standard prompting in most tasks.
This finding suggests that rationale-augmented ensembles provide a reliable approach to improving the final task performance of \textbf{rationale-based few-shot in-context learning}.
Interpretability of model predictions is also enhanced by the presence of generated rationales in the model outputs.
\end{itemize}

\begin{table}[h]
\vspace{-0.1in}
\small
\setlength\tabcolsep{3pt}
    \centering
    \begin{tabular}{c|c|c|c|c|c|c}
    \toprule
    & Input & Output & ANLI R1 / R2 / R3 & e-SNLI & RTE & MNLI-m/mm\\
    \midrule
    Zero-shot CoT \citep{step_by_step} & fixed & greedy & 49.7 / 45.1 / 44.8  & 70.4 & 72.2 & 60.0 / 62.2\\
    \midrule
    Standard-prompting (no-rationale) & fixed & greedy & 69.1 / 55.8 / 55.8 &  85.8 & 84.8 & \textbf{82.7} / \textbf{81.5} \\
    CoT-prompting \citep{wei2022chain} & fixed & greedy & 68.8 / 58.9 / 60.6 & 81.0 & 79.1 & 72.0 / 74.0\\
    \midrule
    \multirow{2}{*}{Prompt-order ensemble} & shuffled & greedy & 72.0 / 60.7 / 61.3  & 84.2 & 78.0 & 74.5 / 75.7 \\
     & shuffled & sampled & \textbf{78.7} / \textbf{64.9} / \textbf{66.0} & \textbf{89.0} & \textbf{84.8} & 80.3 / 81.2\\
    \midrule
    \multirow{2}{*}{Input-rationale ensemble} & sampled & greedy & 70.1 / 60.1 / 61.1 & 87.1 & 79.1 & 73.4 / 75.9 \\
     & sampled & sampled & \textbf{78.3} / \textbf{64.5} / \textbf{64.3} & \textbf{88.8} & \textbf{85.2} & 78.8 / 81.0\\
    \midrule
    Self-consistency \citep{self_consistency} & fixed & sampled & \textbf{78.5} / \textbf{64.5} / \textbf{63.4} & \textbf{88.4} & \textbf{86.3} & 79.5 / 80.5\\
    \bottomrule
    \end{tabular}
    \vspace{0.02in}
    \caption{Performance comparison over \textbf{natural language inference} tasks, on PaLM-540B.}
    \label{tab:performance_1}
% \vspace{-0.1in}
\end{table}

We explain these experiments in more detail.
Table~\ref{tab:performance_1} shows the results obtained across a range of natural language inference tasks.
One can see that the three rationale-augmented ensembling strategies (``output-sampled'') all achieve significantly higher accuracy than chain-of-thought prompting with human-written rationales \citep{wei2022chain}.
On e-SNLI, RTE, and MNLI, the chain-of-thought approach produces worse performance than standard prompting, while rationale-augmented ensembling is able to boost the performance significantly, outperforming chain-of-thought prompting in every case, and outperforming standard prompting in all cases except MNLI. 

\begin{table}[h]
\vspace{-0.1in}
\small
\setlength\tabcolsep{2pt}
    \centering
    \begin{tabular}{c|c|c|c|c|c|c}
    \toprule
    & Input & Output & \makecell{BoolQ \\(q only)} & \makecell{BoolQ \\(w/ passage)} & \makecell{HotpotQA\\ (q only, EM/F1)} & \makecell{OpenBookQA \\(q only)} \\
    \midrule
    Zero-shot CoT \citep{step_by_step}& fixed & greedy & 55.4 & 71.7 & 17.1 / 23.0 & 67.6\\
    \midrule
    Standard-prompting (no-rationale)& fixed & greedy & 71.3 & 89.7 & 27.1 / 36.8 & 84.4 \\
    CoT-prompting \citep{wei2022chain} & fixed & greedy & 74.2 & 85.4 & 28.9 / 39.8 & 86.4 \\
    \midrule
    \multirow{2}{*}{Prompt-order ensemble} & shuffled & greedy & 73.3 & 87.4 & 30.3 / 41.3 & 87.6\\
    & shuffled & sampled & \textbf{78.0} & \textbf{91.0} & \textbf{34.7} / \textbf{45.4} & \textbf{91.0}\\
    \midrule
    \multirow{2}{*}{Input-rationale ensemble} & sampled & greedy 
    & 75.0 & 86.5 & 30.9 / 41.7 & 87.4\\
     & sampled & sampled & \textbf{78.6} & \textbf{90.9} & \textbf{32.4 / 43.7} & \textbf{90.0}\\
    \midrule
    Self-consistency \citep{self_consistency} & fixed & sampled & \textbf{78.4} & \textbf{90.6} & \textbf{33.8 / 44.6}& \textbf{90.0}\\
    \bottomrule
    \end{tabular}
    \vspace{0.02in}
    \caption{Performance comparison over \textbf{question answering} tasks on PaLM-540B. For BoolQ we evaluated both closed-book setting (question only, passage not provided) and the setting where passage is provided. For HotpotQA we evaluated the closed-book setting (question only).}
    \label{tab:performance_2}
\vspace{-0.1in}
\end{table}

Similarly, Table~\ref{tab:performance_2} shows the results obtained in four question answering tasks.
For BoolQ, we conducted an evaluation in both the closed-book setting (the model is given a question only, without providing a relevant passage), as well as the setting where both the question and a relevant passage are provided.
In both settings, rationale-augmented ensembling (``output-sampled'') outperforms standard and chain-of-thought prompting. 
For HotpotQA, we conducted the evaluation in the closed-book setting, where rationale-augmented ensembling achieves a higher EM and F1 score compared to both baselines.
An important question for open-domain question answering in the closed-book setting is how to utilize a language model to retrieve relevant knowledge and compose the final answer \citep{roberts-etal-2020-much}.
Our results show that rationale-augmented ensembling is an effective technique for encouraging a language model to retrieve \textit{multiple} relevant facts for a given question.
To illustrate this phenomenon, Table~\ref{tab:open_qa} demonstrates some of the retrieved facts used to yield final answers.

\begin{table}[h]
\vspace{-0.1in}
\scriptsize
    \centering
    \begin{tabular}{p{2.4cm}|p{3.4cm}| p{2.8cm}|p{3.6cm}}
    \toprule
    Task / Question & Golden passage & \multicolumn{2}{p{6.4cm}}{Retrieved facts and composed rationales from the language model (closed-book setting, passage not provided)}\\
    \midrule
        % \textbf{[BoolQ]} is barq's root beer a pepsi product (answer: No) & Barq's, created by Edward Barq..., is owned by the Barq family but \textcolor{teal}{bottled by the Coca-Cola Company}... & Barq's is a brand of root beer \textcolor{blue}{produced by the Coca-Cola company}. The answer is no. & Barq's is a soft drink brand \textcolor{blue}{of the Coca-Cola Company}, and Barq's Root Beer is its flagship product. The answer is no. \\
        % \midrule
        \textbf{[BoolQ]} can u drive in canada with us license (answer: yes) & Persons driving into Canada must have their vehicle's registration document and proof of insurance. & \textcolor{blue}{Yes, you can drive in Canada with a U.S. licence}. You need to have a signed letter from your car rental company. & \textcolor{blue}{49 U.S. states and 6 Canadian provinces have a reciprocal agreement for the recognition of driver licensing}. As long as you have a valid U.S. driver's license, you may drive in Canada. The answer is yes.\\
        \midrule
        \textbf{[HotpotQA]} What government position was held by the woman who portrayed Corliss Archer in the film Kiss and Tell? (answer: Chief of Protocol) & Kiss and Tell is ... \textcolor{teal}{starring then 17-year-old Shirley Temple as Corliss Archer}. $\mid$ Shirley Temple Black... was named \textcolor{teal}{United States ambassador} to Ghana and to Czechoslovakia and also served as \textcolor{teal}{Chief of Protocol of the United States}. & \textcolor{blue}{Shirley Temple Black played Corliss Archer in Kiss and Tell}. \textcolor{blue}{Black was the United States Ambassador} to Ghana and Czechoslovakia. The answer is \textcolor{blue}{Ambassador}. & Corliss Archer was a fictional character. \textcolor{blue}{Actress Shirley Temple portrayed Corliss Archer in the film Kiss and Tell}. In 1967, \textcolor{blue}{Shirley Temple became the first female Chief of Protocol} in the United States. The answer is \textcolor{blue}{Chief of Protocol}.\\
        \bottomrule
    \end{tabular}
    \vspace{0.01in}
    \caption{Examples of how the language model retrieves multiple relevant facts and composes rationales for open-domain question-answering in the closed-book setting.}
    \label{tab:open_qa}
\vspace{-0.1in}
\end{table}

\begin{table}[h]
\vspace{-0.1in}
\small
\setlength\tabcolsep{2pt}
    \centering
    \begin{tabular}{c|c|c|c|c|c|c|c|c}
    \toprule
    & Input & Output & WiC & SST-2 & QQP & ARC-e & ARC-c & GSM8K \\
    \midrule
    Zero-shot CoT \citep{step_by_step} & fixed & greedy & 54.1 & 76.8 & 55.8 & 87.0 & 79.6 & 43.0 \\
    \midrule
    Standard-prompting (no-rationale) & fixed & greedy & 67.6  & \textbf{94.6} & \textbf{84.1} & 95.9 & 87.1 & 17.9\\
    CoT-prompting \citep{wei2022chain} & fixed & greedy & 65.2 & 87.8 & 75.6 & 95.3 & 85.2 & 56.5 \\
    \midrule
    \multirow{2}{*}{Prompt-order ensemble} & shuffled & greedy & 62.1 & 88.1 & 76.6 & 94.5 & 85.6 & 59.6\\
    & shuffled & sampled & 62.5 & 91.2 & 80.9 & \textbf{96.4} & \textbf{88.5} & \textbf{75.4} \\
    \midrule
    \multirow{2}{*}{Input-rationale ensemble} & sampled & greedy & 66.5 / \textbf{72.1} & 92.3 & 76.6 & 95.5 & 86.6 & 58.9 \\
    & sampled & sampled & 65.2 / \textbf{70.8}  & 93.1 & 81.2 & \textbf{96.7} & \textbf{88.6} & \textbf{73.8}\\
    \midrule
    Self-consistency \citep{self_consistency} & fixed & sampled & 66.9 & 91.1 & 78.9 & \textbf{96.4} & \textbf{88.7} & \textbf{74.4} \\
    \bottomrule
    \end{tabular}
    \vspace{0.02in}
    \caption{Performance comparison over other common NLP tasks, on PaLM-540B.}
    \label{tab:performance_3}
\vspace{-0.2in}
\end{table}

Finally, Table~\ref{tab:performance_3} provides results for other common natural language processing tasks.
Interestingly, for tasks that do not require explicit intermediate steps, such as SST-2 and QQP, adding manual rationales to prompts can degrade performance significantly.
Yet, in these cases, rationale-augmented ensembles (``output-sampled'') are able to significantly close the gap.
For WiC, ARC-easy/challenge and GSM8K, rationale-augmented ensembling outperforms both standard and chain-of-thought prompting by a large margin.
Here, for WiC, we evaluated an alternative variant of the input-rationale ensemble: instead of replacing one rationale in each prompt, we replace every original rationales by a generated one in each prompt.
This variant generally yields similar or slightly worse performance compared to replacing one rationale at a time, but on the WiC task we observed a performance improvement (70.8\% versus 65.2\% when only one rationale is replaced), which indicates that this task might require greater rationale diversity to support strong task performance.

\subsection{Results on GPT-3}
\label{sec:gpt3}

To control for the effects of the language model and aid reproducibility, we repeat the above studies with the publicly available GPT-3 model \citep{brown2020language,instructGPT}. 
Once again, we find similar outcomes where rationale-augmented ensembling robustly improves performance across natural language tasks.
Here we use the code-davinci-002 engine \citep{chen2021evaluating}, which has been observed to yield slightly better performance than text-davinci-002.
The results of this study are given in Table~\ref{tab:performance_gpt}, showing that rationale-augmented ensembles with GPT-3 obtain similar improvements to those obtained with PaLM-540B above. 
Once again, human-written rationales in few-shot learning can sometimes degrade performance compared to standard prompting (e.g., on RTE, OpenBookQA, WiC, ARC-challenge), while rationale-augmented ensembling with sampling in the output space (``output-sampled'') reliably improves performance over both baselines.
Similarly, for WiC, introducing greater diversity in sampled rationales improves performance (67.6\%) compared to sampling a single rationale for each prompt (57.4\%).
These results reinforce the finding that the improvements are robust to the specific language model, provided it is of sufficient size/quality.

\begin{table}[h]
\vspace{-0.1in}
\small
\setlength\tabcolsep{3pt}
    \centering
    \begin{tabular}{c|c|c|c|c|c|c|c}
    \toprule
    & Input & Output & RTE & BoolQ & OpenBookQA & WiC & ARC-c \\
    \midrule
    Standard-prompting (no-rationale) & fixed & greedy & 85.2 & 69.9 & 81.4 & 65.5 & 85.9\\
    CoT-prompting \citep{wei2022chain} & fixed & greedy & 84.1 & 73.5 & 80.4 & 55.5 & 83.6\\
    \midrule
    \multirow{2}{*}{Prompt-order ensemble} & shuffled & greedy 
    & 83.0 & 74.2 & 83.4 & 56.4 & 84.0\\
    & shuffled & sampled & \textbf{88.8} & \textbf{78.5} & \textbf{87.8} & 56.7 & \textbf{88.2}\\
    \midrule
    \multirow{2}{*}{Input-rationale ensemble} & sampled & greedy & 85.2 & 75.0 & 85.4 & 57.1 / \textbf{68.0} & 84.7 \\
    & sampled & sampled & \textbf{87.4} & \textbf{78.4} & \textbf{87.0} & 57.4 / \textbf{67.6} & \textbf{87.6}\\
    \midrule
    Self-consistency \citep{self_consistency} & fixed & sampled & 85.6 & \textbf{78.2} & \textbf{88.4} & 55.6 & \textbf{87.5}\\
    \bottomrule
    \end{tabular}
    \vspace{0.02in}
    \caption{Performance comparison on GPT-3 (code-davinci-002 engine).}
    \label{tab:performance_gpt}
\vspace{-0.2in}
\end{table}

\subsection{Additional Studies}

\paragraph{Effect of $K$ in $K$-shot in-context learning.}
In Table~\ref{tab:ablation_shot}, we provide an ablation study that examines the effect of choosing different $K$ in $K$-shot in-context learning.
While increasing the number of exemplars $K$ generally improves performance, rationale-augmented ensembling robustly improves performance over standard and chain-of-thought prompting for all values of $K$.

\begin{table}[h]
\vspace{-0.1in}
\small
\setlength\tabcolsep{5pt}
    \centering
    \begin{tabular}{c|c|c|c|c|c||c|c}
    \toprule
    & Input & Output & 3-shot & 6-shot/T-1 & 9-shot & T-2 & T-3\\
    \midrule
    Standard-prompting (no-rationale) & fixed & greedy & 67.9 & 69.1 & 69.3 &  66.1 & 66.4 \\
    CoT-prompting \citep{wei2022chain} & fixed & greedy & 71.6 & 68.8 & 72.2 &  67.9 & 68.3\\
    \midrule
    Prompt-order ensemble & shuffled & sampled & 76.0 & 78.7 & 80.1 & 78.4 & 75.6\\
    Input-rationale ensemble & sampled & sampled & 76.1 & 78.3 & 78.4 & 77.8 & 76.0\\
    Self-consistency \citep{self_consistency}& fixed & sampled & 77.9 & 78.5 & 78.7 & 76.6 & 76.9\\
    \bottomrule
    \end{tabular}
    \vspace{0.02in}
    \caption{Performance comparison on ANLI-R1 using PaLM-540B, with (1) varying $K$ (3, 6, 9) in $K$-shot learning; and (2) using different templates/verbalizers (T-1, T-2, T-3), fixing $K=6$.}
    \label{tab:ablation_shot}
\vspace{-0.2in}
\end{table}

\paragraph{Effect of templates and verbalizers.}
We also investigate whether rationale-augmented ensembling is robust to different templates or verbalizers, since previous work has shown that templates or verbalizers can have a significant effect on final performance \citep{bach2022promptsource}. 
Here we choose three alternative templates from PromptSource%
\footnote{\url{https://github.com/bigscience-workshop/promptsource}}
for the NLI task, as follows:
\begin{itemize}[leftmargin=0.6cm]
    \item Template-1: \textit{Premise:\textcolor{blue} {\textbackslash\textbackslash}\{premise\}"\textcolor{blue} {\textbackslash\textbackslash}Based on this premise, can we conclude the hypothesis "\{hypothesis\}" ... is true?{\textbackslash\textbackslash}{options}}
    \item Template-2: \textit{"\{premise\}"\textcolor{blue} {\textbackslash\textbackslash}Does it follow that "\{hypothesis\}"?\textcolor{blue} {\textbackslash\textbackslash}{options}}
    \item Template-3: \textit{Suppose "{premise}"\textcolor{blue} {\textbackslash\textbackslash}Can we infer that "{hypothesis}"?\textcolor{blue} {\textbackslash\textbackslash}{options}}
\end{itemize}

The results in Table~\ref{tab:ablation_shot} reveal that, although different templates can induce variable performance, rationale-augmented ensembling outperforms standard and chain-of-thought prompting under all three templates.

\paragraph{Effect of using existing explanations vs newly-written ones in the prompts.}
To control for the bias of manually written rationales, we also investigate performance on the e-SNLI dataset using crowd-sourced rationales \citep{esnli}.
As shown in Table~\ref{tab:performance_1}, the improvement of rationale-augmented ensemble appears to be stable regardless of whether the rationales are crowd-sourced or author-supplied.

Note that in this paper, we focus on the role of ``rationales'', and conduct the studies in a manner that fixes other factors that might affect task performance.
Due to the large performance variance across alternative set-ups, it is clear that a rigorous evaluation of few-shot in-context learning requires the specification of all these factors, including (1) the exact prompts used, including the specific exemplars, templates/verbalizers, instructions, or rationales/explanations used; and (2) the exact prompt order and the number of exemplars $K$ used.

\section{Related work}

% There are a number of lines of related work leading to this study.

\paragraph{Rationalization and interpretability in NLP.}
One relevant line of work tries to improve rationalization and interpretability in natural language processing models, for example, by extracting rationales using task-specific approaches \citep{xu-etal-2021-exploiting-reasoning,Asai2020Learning,DBLP:journals/corr/abs-1910-02610}. 
In the supervised learning setting, one typically fine-tunes a model using human-annotated rationales as training data \citep{zaidan-etal-2007-using,ling-etal-2017-program,wt5,cobbe2021training}.
% \cite{scratchpad} propose to add intermediate computation steps for tasks such as integer addition and program synthesis and show it can further improve model's performance in fine-tuning.
\cite{star} propose to use prompting to augment a training dataset with rationales, then fine-tune a language model using this dataset to further improve reasoning ability. 
\cite{better_reasoner} propose to sample ``diverse'' prompts from the training set augmented by rationales, plus an additional voting verifier to improve model performance on reasoning tasks.
However, the use of an additional training set is closer to the fine-tuning setting rather than the few-shot setting.
Compared to these approaches, rationale-augmented ensembles focus more on the few-shot setting, where there is no additional training or fine-tuning, hence no human annotation nor training/development datasets are required.

Recent work has also considered \textit{prompting} language models with human-written rationales to further improve performance, such as \citep{wei2022chain,step_by_step,self_consistency,maieutic_prompting}.
\cite{explanation_deepmind} show that hand-tuned explanations can improve task performance substantially.
By contrast, rationale-augmented ensembling requires no hand-tuning on rationales.
Instead, we leverage the language model to automatically sample rationales to overcome the sub-optimality of manually provided rationales.

\paragraph{Prompt optimization and ensembles in language models.}
Previous work has shown that the prompt order \citep{Lu2021FantasticallyOP}, how each task is verbalized \citep{bach2022promptsource}, and the distribution of labels in the prompts \citep{pmlr-v139-zhao21c} can all affect final task performance. 
In this paper, we find that, when shifting from the paradigm of (input $\rightarrow$ output) pairs to (input, \textit{rationale} $\rightarrow$ output) pairs, there is also a large variance in the final task performance when the \textit{rationales} used in the prompts differ.
Recent work has also proposed ways to further improve a model's reasoning ability under specific constraints. 
For example, when the final label is binary, \cite{maieutic_prompting} induce a tree of explanations, then use an SAT solver and an NLI verifier to infer the satisfiability of each explanation.
For commonsense reasoning tasks, \cite{liu-etal-2022-generated} generate relevant knowledge as additional inputs to the model, to improve the performance. 
Another line of work proposes to better retrieve prompts closer to the target question to further improve task performance \citep{liu-etal-2022-makes,learning_to_retrieve}.

\paragraph{Learn to execute programs with intermediate computations.}
Although much of the work on rationales has come from the natural language processing literature, there has been growing interest in similar mechanisms in the area of program synthesis.
\cite{scratchpad} use pretrained language models to execute a program by predicting the intermediate states of a program behaviour line-by-line.
This work shows that eliciting step-by-step reasoning described by a formal language can dramatically improve the execution prediction accuracy.
Other recent work \citep{pi2022reasoning} pre-trains language models as program executors and shows that this can improve reasoning task performance.

% \section{Conclusion and Discussion}
% In this paper, we unified the framework of rationale-augmented ensembles, and identify the key component as rationale sampling in the output space for a better final task performance. By sampling diverse rationales and ensembling the results, we show that this framework can reliably outperform standard prompting and rationale-based few-shot prompting, across a wide range of NLP tasks over multiple language models.
% Rationale-augmented ensemble is a reliable way to shift from the paradigm of (input $\rightarrow$ output) pairs to (input, \textit{rationale} $\rightarrow$ output) pairs, as well as towards more accurate and interpretable natural language understanding.

% On the other hand, although the framework avoids the sub-optimality of the human-written rationales to a large extent, some human-written seed rationales are still required and might bias the generation of the output rationales. We noticed that in the few-shot in-context learning setting, the patterns of the rationales written could largely affect the model's generated rationales, hence some diversity in the seed rationales could help improve the diversity in the outputs.
% We also call for additional research on a deeper understanding of how models respond differently over slight variations on the few-shot exemplars and more automatic approaches to generate better prompts for a given task.

\section{Conclusion}
In this paper, we have presented a unified framework for rationale-augmented ensembles, and found that rationale sampling in the output space is a key component for achieving improved performance in natural language processing tasks. 
By sampling diverse rationales and ensembling the results, we have shown that rational-ensembling methods in the proposed framework can reliably outperform standard prompting and rationale-based few-shot prompting, across a wide range of natural language tasks and alternative language models.
Overall, rationale-augmented ensembling appears to be a reliable way to shift from the paradigm of (input $\rightarrow$ output) pairs to (input, \textit{rationale} $\rightarrow$ output) pairs to achieve more accurate and interpretable natural language processing.

Although the proposed framework mitigates sensitivity to human-written rationales, some human-written seed rationales are still required, which could still bias generation of output rationales.
We have observed that patterns expressed in the written rationales can affect a model's generated rationales.
For example, if all seed rationales are written in a similar style, like ``The first...the second...'', subsequently generated rationales will tend to follow the same pattern. 
Therefore, some diversity in seed rationales still appears to be important for inducing sufficient diversity in generated rationales.

Overall, through this study, we hope to motivate more research on understanding how language models respond differently to variations in few-shot exemplars, which can lead to the development of more robust and autonomous approaches for generating effective prompts for a given target task.

\bibliographystyle{icml2022}
\bibliography{main}

%%%%%%%%%%%%%%%%%%%%%%%%%%%%%%%%%%%%%%%%%%%%%%%%%%%%%%%%%%%%

%%%%%%%%%%%%%%%%%%%%%%%%%%%%%%%%%%%%%%%%%%%%%%%%%%%%%%%%%%%%

%%%%%%%%%%%%%%%%%%%%%%%%%%%%%%%%%%%%%%%%%%%%%%%%%%%%%%%%%%%%
\clearpage
\appendix

\section{Appendix}

\subsection{Full Sets of Prompts}
\label{sec:appendix-prompt}

\begin{table}[h]
    \caption{
    Few-shot exemplars for RTE.
    }
    \centering
    \small
    \begin{tabular}{p{14cm}}
        \toprule
        Premise:\\
        "No Weapons of Mass Destruction Found in Iraq Yet."\\
        Based on this premise, can we conclude the hypothesis "Weapons of Mass Destruction Found in Iraq." is true?\\
        \vspace{-1mm}
        \textbf{A:} "No Weapons of Mass Destruction Found" contradicts "Weapons of Mass Destruction Found". The answer is no.\\
        \vspace{0mm}
        Premise:\\
        "A place of sorrow, after Pope John Paul II died, became a place of celebration, as Roman Catholic faithful gathered in downtown Chicago to mark the installation of new Pope Benedict XVI."\\
        Based on this premise, can we conclude the hypothesis "Pope Benedict XVI is the new leader of the Roman Catholic Church." is true?'\\
        \vspace{-1mm}
        \textbf{A:} "installation of new Pope Benedict XVI." means "Pope Benedict XVI is the new leader". The answer is yes. \\
        \vspace{0mm}
        Premise:\\
        "A man is due in court later charged with the murder 26 years ago of a teenager whose case was the first to be featured on BBC One's Crimewatch. Colette Aram, 16, was walking to her boyfriend's house in Keyworth, Nottinghamshire, on 30 October 1983 when she disappeared. Her body was later found in a field close to her home. Paul Stewart Hutchinson, 50, has been charged with murder and is due before Nottingham magistrates later."\\
        Based on this premise, can we conclude the hypothesis "Paul Stewart Hutchinson is accused of having stabbed a girl." is true?\\
        \vspace{-1mm}
        \textbf{A:} The premise does not say Paul Stewart Hutchinson "stabbed" this girl. The answer is no.\\
        \vspace{0mm}
        Premise:\\
        "Herceptin was already approved to treat the sickest breast cancer patients, and the company said, Monday, it will discuss with federal regulators the possibility of prescribing the drug for more breast cancer patients."\\
        Based on this premise, can we conclude the hypothesis "Herceptin can be used to treat breast cancer." is true?\\
        \vspace{-1mm}
        \textbf{A:} "Herceptin was approved to treat breast cancer" implies that "Herceptin can be used to treat breast cancer". The answer is yes.\\
        \bottomrule
    \end{tabular}
    \label{tab:prompt-rte}
\end{table}

\begin{table}[h]
    \caption{
    Few-shot exemplars for ARC easy/challenge.
    }
    \centering
    \small
    \begin{tabular}{p{14cm}}
        \toprule
        \textbf{Q:} George wants to warm his hands quickly by rubbing them. Which skin surface will produce the most heat? (a) dry palms. (b) wet palms. (c) palms covered with oil. (d) palms covered with lotion.\\
        \vspace{-1mm}
        \textbf{A:} Dry surfaces will more likely cause more friction via rubbing than other smoother surfaces, hence dry palms will produce the most heat. The answer is (a).\\
        \vspace{0mm}
        \textbf{Q:} Which factor will most likely cause a person to develop a fever? (a) a leg muscle relaxing after exercise. (b) a bacterial population in the bloodstream. (c) several viral particles on the skin. (d) carbohydrates being digested in the stomach.\\
        \vspace{-1mm}
        \textbf{A:} Option (b), bacterial population is the most likely cause for a person developing fever. The answer is (b). \\
        \vspace{0mm}
        \textbf{Q:} Which change in the state of water particles causes the particles to become arranged in a fixed position? (a) boiling. (b) melting. (c) freezing. (d) evaporating.\\
        \vspace{-1mm}
        \textbf{A:} When water is freezed, the particles are arranged in a fixed position; the particles are still moving for all other options. The answer is (c).\\
        \vspace{0mm}
        \textbf{Q:} When a switch is used in an electrical circuit, the switch can (a) cause the charge to build. (b) increase and decrease the voltage. (c) cause the current to change direction. (d) stop and start the flow of current.\\
        \vspace{-1mm}
        \textbf{A:} The function of a switch is to start and stop the flow of a current. The answer is (d).\\
        \bottomrule
    \end{tabular}
    \label{tab:prompt-arc}
\end{table}

\begin{table}[h]
    \caption{
    Few-shot exemplars for NLI tasks, including ANLI and MNLI.
    }
    \centering
    \small
    \begin{tabular}{p{14cm}}
        \toprule
        Premise:\\
        "Conceptually cream skimming has two basic dimensions - product and geography."\\
        Based on this premise, can we conclude the hypothesis "Product and geography are what make cream skimming work." is true?\\
        OPTIONS:\\
        - yes\\
        - no\\
        - it is not possible to tell\\
        \vspace{-1mm}
        \textbf{A:} Based on "cream skimming has two basic dimensions" we can't infer that these two dimensions are what make cream skimming work. The answer is it is not possible to tell.\\
        \vspace{0mm}
        Premise:\\
        "One of our member will carry out your instructions minutely."\\
        Based on this premise, can we conclude the hypothesis "A member of my team will execute your orders with immense precision." is true?\\
        OPTIONS:\\
        - yes\\
        - no\\
        - it is not possible to tell\\
        \vspace{-1mm}
        \textbf{A:} "one of" means the same as "a member of", "carry out" means the same as "execute", and "minutely" means the same as "immense precision". The answer is yes.\\
        \vspace{0mm}
        Premise:\\
        "Fun for adults and children."\\
        Based on this premise, can we conclude the hypothesis "Fun for only children." is true?\\
        OPTIONS:\\
        - yes\\
        - no\\
        - it is not possible to tell\\
        \vspace{-1mm}
        \textbf{A:} "adults and children" contradicts "only children". The answer is no.\\
        \vspace{0mm}
        Premise:\\
        "He turned and smiled at Vrenna."\\
        Based on this premise, can we conclude the hypothesis "He smiled at Vrenna who was walking slowly behind him with her mother." is true?\\
        OPTIONS:\\
        - yes\\
        - no\\
        - it is not possible to tell\\
        \vspace{-1mm}
        \textbf{A:} the premise does not say anything about "Vrenna was walking". The answer is it is not possible to tell.\\
        \vspace{0mm}
        Premise:\\
        "well you see that on television also"\\
        Based on this premise, can we conclude the hypothesis "You can see that on television, as well." is true?\\
        OPTIONS:\\
        - yes\\
        - no\\
        - it is not possible to tell\\
        \vspace{-1mm}
        \textbf{A:} "also" and "as well" mean the same thing. The answer is yes.\\
        \vspace{0mm}
        Premise:\\
        "Vrenna and I both fought him and he nearly took us."\\
        Based on this premise, can we conclude the hypothesis "Neither Vrenna nor myself have ever fought him." is true?\\
        OPTIONS:\\
        - yes\\
        - no\\
        - it is not possible to tell\\
        \vspace{-1mm}
        \textbf{A:} "Vrenna and I both" contradicts "neither Vrenna nor myself". The answer is no.\\
        \bottomrule
    \end{tabular}
    \label{tab:prompt-nli}
\end{table}

\begin{table}[h]
    \caption{
    Few-shot exemplars for e-SNLI (the rationales are crowd-sourced annotations from the original dataset).
    }
    \centering
    \small
    \begin{tabular}{p{14cm}}
        \toprule
        Premise:\\
"A person on a horse jumps over a broken down airplane."\\
Based on this premise, can we conclude the hypothesis "A person is training his horse for a competition." is true?\\
OPTIONS:\\
- yes\\
- no\\
- it is not possible to tell\\
\textbf{A:} The person is not necessarily training his horse. The answer is it is not possible to tell.\\
\vspace{0mm}
Premise:\\
"A person on a horse jumps over a broken down airplane."\\
Based on this premise, can we conclude the hypothesis "A person is at a diner, ordering an omelette." is true?\\
OPTIONS:\\
- yes\\
- no\\
- it is not possible to tell\\
\textbf{A:} One jumping horse cannot be in a diner ordering food. The answer is no.\\
\vspace{0mm}
Premise:\\
"A person on a horse jumps over a broken down airplane."\\
Based on this premise, can we conclude the hypothesis "A person is outdoors, on a horse." is true?\\
OPTIONS:\\
- yes\\
- no\\
- it is not possible to tell\\
\textbf{A:} A broken down airplane is outdoors. The answer is yes.\\
\vspace{0mm}
Premise:\\
"Children smiling and waving at camera."\\
Based on this premise, can we conclude the hypothesis "They are smiling at their parents." is true?\\
OPTIONS:\\
- yes\\
- no\\
- it is not possible to tell\\
\textbf{A:} Just because they are smiling and waving at a camera does not imply their parents or anyone is anyone behind it. The answer is it is not possible to tell.\\
\vspace{0mm}
Premise:\\
"Children smiling and waving at camera."\\
Based on this premise, can we conclude the hypothesis "The kids are frowning." is true?
OPTIONS:\\
- yes\\
- no\\
- it is not possible to tell\\
\textbf{A:} One cannot be smiling and frowning at the same time. The answer is no.\\
\vspace{0mm}
Premise:\\
"Children smiling and waving at camera."\\
Based on this premise, can we conclude the hypothesis "There are children present." is true?\\
OPTIONS:\\
- yes\\
- no\\
- it is not possible to tell\\
\textbf{A:}The children must be present to see them smiling and waving. The answer is yes.\\
        \bottomrule
    \end{tabular}
    \label{tab:prompt-esnli}
\end{table}

\begin{table}[h]
    \caption{
    Few-shot exemplars for BoolQ (closed-book setting).
    }
    \centering
    \small
    \begin{tabular}{p{14cm}}
        \toprule
        \textbf{Q:} does system of a down have 2 singers?\\
        \vspace{-1mm}
        \textbf{A:} System of a Down currently consists of Serj Tankian, Daron Malakian, Shavo Odadjian and John Dolmayan. Serj and Daron do vocals, so the band does have two singers. The answer is yes.\\
        \vspace{0mm}
        \textbf{Q:} do iran and afghanistan speak the same language?\\
        \vspace{-1mm}
        \textbf{A:} Iran and Afghanistan both speak the Indo-European language Persian. The answer is yes. \\
        \vspace{0mm}
        \textbf{Q:} is a cello and a bass the same thing?\\
        \vspace{-1mm}
        \textbf{A:} The cello is played sitting down with the instrument between the knees, whereas the double bass is played standing or sitting on a stool. The answer is no.\\
        \vspace{0mm}
        \textbf{Q:} can you use oyster card at epsom station?\\
        \vspace{-1mm}
        \textbf{A:} Epsom railway station serves the town of Epsom in Surrey and is not in the London Oyster card zone. The answer is no.\\
        \bottomrule
    \end{tabular}
    \label{tab:prompt-boolq-cb}
\end{table}

\begin{table}[h]
    \caption{
    Few-shot exemplars for BoolQ (with passage provided).
    }
    \centering
    \small
    \begin{tabular}{p{14cm}}
        \toprule
        \textbf{Q:} System of a Down, sometimes shortened to System and abbreviated as SOAD, is an Armenian-American heavy metal band from Glendale, California, formed in 1994. The band currently consists of Serj Tankian (lead vocals, keyboards), Daron Malakian (vocals, guitar), Shavo Odadjian (bass, backing vocals) and John Dolmayan (drums).\\
        Based on the above text, does system of a down have 2 singers?\\
        \vspace{-1mm}
        \textbf{A:} System of a Down currently consists of Serj Tankian, Daron Malakian, Shavo Odadjian and John Dolmayan. Serj and Daron do vocals, so the band does have two singers. The answer is yes.\\
        \vspace{0mm}
        \textbf{Q:} Persian, also known by its endonym Farsi, is one of the Western Iranian languages within the Indo-Iranian branch of the Indo-European language family. It is primarily spoken in Iran, Afghanistan, and Tajikistan, and some other regions which historically were Persianate societies and considered part of Greater Iran.\\
        Based on the above text, do iran and afghanistan speak the same language?'\\
        \vspace{-1mm}
        \textbf{A:} Iran and Afghanistan both speak the Indo-European language Persian. The answer is yes. \\
        \vspace{0mm}
        \textbf{Q:} Both the violin and viola are played under the jaw. The viola, being the larger of the two instruments, has a playing range that reaches a perfect fifth below the violin's. The cello is played sitting down with the instrument between the knees, and its playing range reaches an octave below the viola's. The double bass is played standing or sitting on a stool, with a range that typically reaches a minor sixth, an octave or a ninth below the cello's.\\
        Based on the above text, is a cello and a bass the same thing?\\
        \vspace{-1mm}
        \textbf{A:} The cello is played sitting down with the instrument between the knees, whereas the double bass is played standing or sitting on a stool. The answer is no.\\
        \vspace{0mm}
        \textbf{Q:} Epsom railway station serves the town of Epsom in Surrey. It is located off Waterloo Road and is less than two minutes' walk from the High Street. It is not in the London Oyster card zone unlike Epsom Downs or Tattenham Corner stations. The station building was replaced in 2012/2013 with a new building with apartments above the station.\\
        Based on the above text, can you use oyster card at epsom station?\\
        \vspace{-1mm}
        \textbf{A:} Epsom railway station serves the town of Epsom in Surrey and is not in the London Oyster card zone. The answer is no.\\
        \bottomrule
    \end{tabular}
    \label{tab:prompt-boolq}
\end{table}

\begin{table}[h]
    \caption{
    Few-shot exemplars for HotpotQA (closed-book setting).
    }
    \centering
    \small
    \begin{tabular}{p{14cm}}
        \toprule
        \textbf{Q:} Which magazine was started first Arthur's Magazine or First for Women?\\
        \vspace{-1mm}
        \textbf{A:} Arthur's Magazine started in 1844. First for Women started in 1989. So Arthur's Magazine was started first. The answer is Arthur's Magazine.\\
        \vspace{0mm}
        \textbf{Q:} The Oberoi family is part of a hotel company that has a head office in what city?\\
        \vspace{-1mm}
        \textbf{A:} The Oberoi family is part of the hotel company called The Oberoi Group. The Oberoi Group has its head office in Delhi. The answer is Delhi. \\
        \vspace{0mm}
        \textbf{Q:} What nationality was James Henry Miller's wife?\\
        \vspace{-1mm}
        \textbf{A:} James Henry Miller's wife is June Miller. June Miller is an American. The answer is American.\\
        \vspace{0mm}
        \textbf{Q:} The Dutch-Belgian television series that "House of Anubis" was based on first aired in what year?\\
        \vspace{-1mm}
        \textbf{A:} "House of Anubis" is based on the Dutch–Belgian television series Het Huis Anubis. Het Huis Anubis is first aired in September 2006. The answer is 2006.\\
        \bottomrule
    \end{tabular}
    \label{tab:prompt-hotpotqa}
\end{table}

\begin{table}[h]
    \caption{
    Few-shot exemplars for OpenBookQA.
    }
    \centering
    \small
    \begin{tabular}{p{14cm}}
        \toprule
        \textbf{Q:} Poison causes harm to which of the following? (a) a Tree (b) a robot (c) a house (d) a car\\
        \vspace{-1mm}
        \textbf{A:} Poison will harm living things, only a tree is a living thing. The answer is (a).\\
        \vspace{0mm}
        \textbf{Q:} As you look deeper into a Marbel you can see (a) the future (b) minut defects (c) colors (d) the other side\\
        \vspace{-1mm}
        \textbf{A:} Marbel is not transparent, so you can not see the other side. Marbel does not necessarily have multiple colors. You will see minut defects. The answer is (b). \\
        \vspace{0mm}
        \textbf{Q:} When food is reduced in the stomach (a) the mind needs time to digest (b) take a second to digest what I said (c) nutrients are being deconstructed (d) reader's digest is a body of works\\
        \vspace{-1mm}
        \textbf{A:} The food is being deconstructed in the stomach during digestion. The answer is (c).\\
        \vspace{0mm}
        \textbf{Q:} The sun is responsible for (a) puppies learning new tricks (b) children growing up and getting old (c) flowers wilting in a vase (d) plants sprouting, blooming and wilting\\
        \vspace{-1mm}
        \textbf{A:} The sun can affect the growing of living things, like plants. The answer is (d).\\
        \bottomrule
    \end{tabular}
    \label{tab:prompt-obqa}
\end{table}

\begin{table}[h]
    \caption{
    Few-shot exemplars for WiC.
    }
    \centering
    \small
    \begin{tabular}{p{14cm}}
        \toprule
        Do you want to come over to my place later?\\
        A political system with no place for the less prominent groups.\\
        \textbf{Q:} Is the word "place" used in the same way in the two sentences above?\\
        \vspace{-1mm}
        \textbf{A:} The first "place" means "home", the second "place" means "room". The answer is no.\\
        \vspace{0mm}
        Approach a task.\\
        To approach the city.\\
        \textbf{Q:} Is the word "approach" used in the same way in the two sentences above?\\
        \vspace{-1mm}
        \textbf{A:} The first "approach" means "deal with", the second "approach" means "come near". The answer is no. \\
        \vspace{0mm}
        The general ordered the colonel to hold his position at all costs.\\
        Hold the taxi.\\
        \textbf{Q:} Is the word "hold" used in the same way in the two sentences above?\\
        \vspace{-1mm}
        \textbf{A:} Both "hold" mean "keep" or "detain". The answer is yes.\\
        \vspace{0mm}
        We like to summer in the Mediterranean.\\
        We summered in Kashmir.\\
        \textbf{Q:} Is the word "summer" used in the same way in the two sentences above?\\
        \vspace{-1mm}
        \textbf{A:} Both "summer" mean "spend the summer". The answer is yes.\\
        \bottomrule
    \end{tabular}
    \label{tab:prompt-wic}
\end{table}

\begin{table}[h]
    \caption{
    Few-shot exemplars for SST-2.
    }
    \centering
    \small
    \begin{tabular}{p{14cm}}
        \toprule
        What is the sentiment of the following sentence?\\
        "that loves its characters and communicates something rather beautiful about human nature"\\
        \vspace{-1mm}
        \textbf{A:} "loves its characters" indicates positive sentiment. The answer is positive.\\
        \vspace{0mm}
        What is the sentiment of the following sentence?\\
        "hide new secretions from the parental units"\\
        \vspace{-1mm}
        \textbf{A:} If people are hiding something, it means the sentiment is on the negative side. The answer is negative. \\
        \vspace{0mm}
        What is the sentiment of the following sentence?\\
        "the greatest musicians"\\
        \vspace{-1mm}
        \textbf{A:} By saying someone being the "greatest", it means positive sentiment. The answer is positive.\\
        \vspace{0mm}
        What is the sentiment of the following sentence?\\
        "contains no wit , only labored gags"\\
        \vspace{-1mm}
        \textbf{A:} "contains no wit" is clearly a negative sentiment. The answer is negative.\\
        \vspace{0mm}
        What is the sentiment of the following sentence?\\
        "demonstrates that the director of such hollywood blockbusters as patriot games can still turn out a small , personal film with an emotional wallop ."\\
        \vspace{-1mm}
        \textbf{A:} "can still turn out a small , personal film with an emotional wallop ." indicates sentiment on the positive side. The answer is positive.\\
        \vspace{0mm}
        What is the sentiment of the following sentence?\\
        "that 's far too tragic to merit such superficial treatment"\\
        \vspace{-1mm}
        \textbf{A:} "far too tragic" and "to merit such superficial treatment" both mean negative sentiments. The answer is negative.\\
        \bottomrule
    \end{tabular}
    \label{tab:prompt-sst}
\end{table}

\begin{table}[h]
    \caption{
    Few-shot exemplars for QQP.
    }
    \centering
    \small
    \begin{tabular}{p{14cm}}
        \toprule
        Is the following question "What causes stool color to change to yellow?" the same as "What can cause stool to come out as little balls?"\\
        \vspace{-1mm}
        \textbf{A:} "change to yellow" and "come out as little balls" mean different things. The answer is no.\\
        \vspace{0mm}
        Is the following question "What can one do after MBBS?" the same as "What do i do after my MBBS?"\\
        \vspace{-1mm}
        \textbf{A:} Both are asking what can a person do after MBBS. The answer is yes. \\
        \vspace{0mm}
        Is the following question "How is air traffic controlled?" the same as "How do you become an air traffic controller?"\\
        \vspace{-1mm}
        \textbf{A:} "How is air traffic controlled" means differently as "how to become a controller". The answer is no.\\
        \vspace{0mm}
        Is the following question "How do I control my horny emotions?" the same as "How do you control your horniness?"\\
        \vspace{-1mm}
        \textbf{A:} "horny emotions" means the same as "horniness". The answer is yes.\\
        \bottomrule
    \end{tabular}
    \label{tab:prompt-qqp}
\end{table}

\end{document}